\documentclass[10pt,twocolumn,letterpaper]{article}

\usepackage{iccv}
\usepackage{times}
\usepackage{epsfig}
\usepackage{graphicx}
\usepackage{amsmath}
\usepackage{amssymb}
\usepackage{colortbl}
\usepackage{xcolor}
\usepackage{bbding}  %x and ticks
\usepackage{multirow}
\usepackage{caption}
\usepackage{xspace}
\usepackage[breaklinks=true,bookmarks=false]{hyperref}
\hypersetup{colorlinks}
\definecolor{mygray}{gray}{.9}
\definecolor{mygray2}{gray}{.8}
\newlength\savewidth\newcommand\shline{\noalign{\global\savewidth\arrayrulewidth
  \global\arrayrulewidth 1pt}\hline\noalign{\global\arrayrulewidth\savewidth}}

\def\ie{\textit{i.e}\onedot} 
 
 \def\vs{\textit{vs}\onedot}

\makeatother

\newcommand{\app}{\raise.17ex\hbox{$\scriptstyle\sim$}}

\iccvfinalcopy % *** Uncomment this line for the final submission

\makeatletter\renewcommand\paragraph{\@startsection{paragraph}{4}{\z@}
  {.3em \@plus1ex \@minus.2ex}{-.5em}{\normalfont\normalsize\bfseries}}\makeatother

\makeatletter
\let\@oldmaketitle\@maketitle% Store \@maketitle
\renewcommand{\@maketitle}{\@oldmaketitle% Update \@maketitle to insert...
     \centering
     \vspace{-2.2em}
     \includegraphics[width=.82\linewidth]{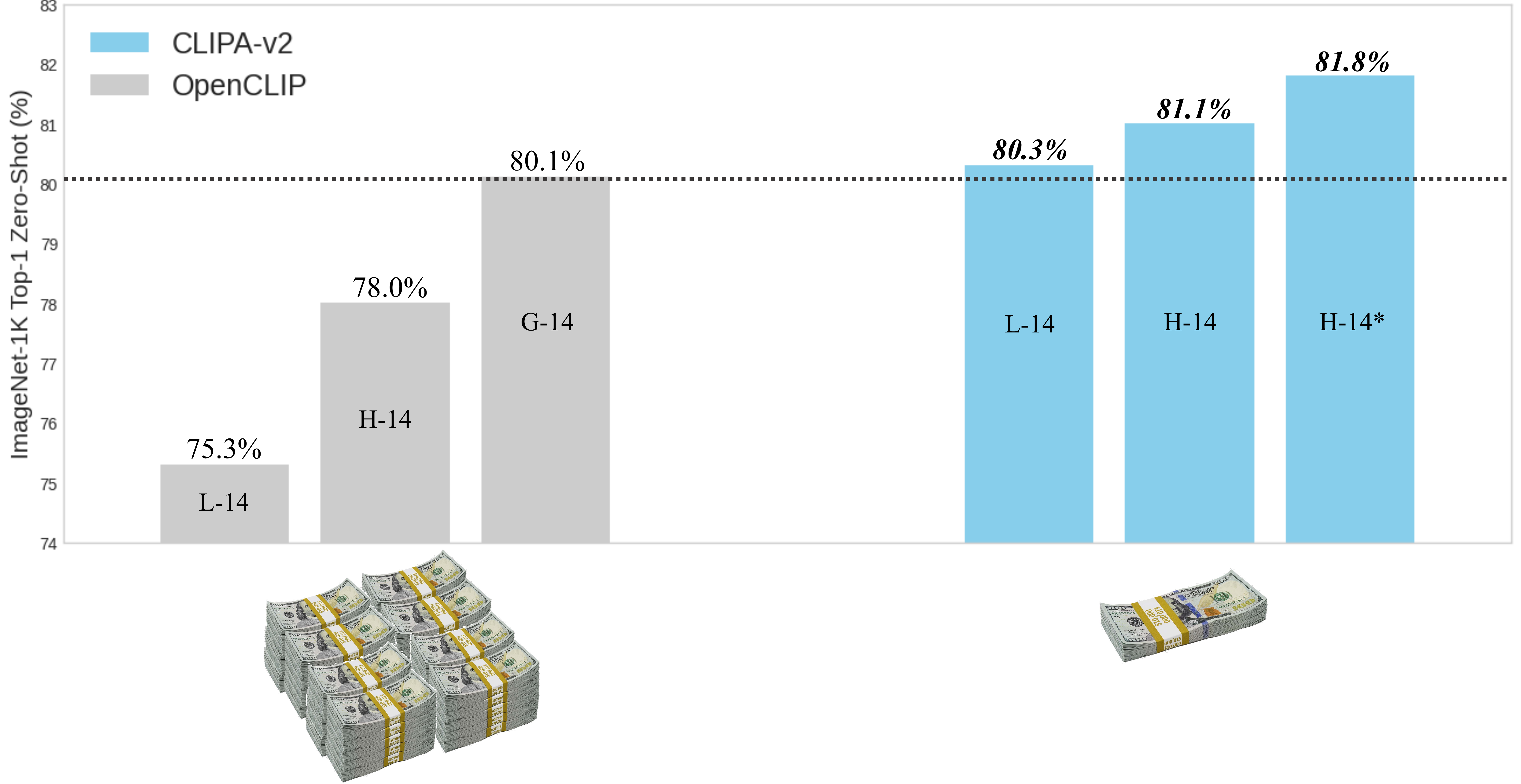}
     \vspace{-.7em}
    \captionof{figure}{Compared to OpenCLIP~\cite{openclip}, our CLIPA-v2 models achieve higher performance with lower training cost.}
    \label{fig:teaser}
    \vspace{-.3em}
    \bigskip}% ... an image
\makeatother

\def\iccvPaperID{****} % *** Enter the ICCV Paper ID here

\ificcvfinal\fi

\begin{document}

%%%%%%%%% TITLE
\title{CLIPA-v2: Scaling CLIP Training with 81.1\% Zero-shot ImageNet Accuracy within a \textdollar10,000 Budget; An Extra \textdollar4,000 Unlocks 81.8\% Accuracy \vspace{-.55em}}

\author{
Xianhang Li\textsuperscript{*} \quad
Zeyu Wang\textsuperscript{*} \quad
Cihang Xie \vspace{.3em}
\\
\small $^{*}$equal technical contribution \vspace{.3em} \\
UC Santa Cruz  \\
\small \url{https://github.com/UCSC-VLAA/CLIPA}
}

\maketitle
\ificcvfinal\thispagestyle{empty}\fi

%%%%%%%%% ABSTRACT
\begin{abstract}
The recent work CLIPA \cite{li2023inverse} presents an \textit{inverse scaling law} for CLIP training --- whereby the larger the image/text encoders used, the shorter the sequence length of image/text tokens that can be applied in training. This finding enables us to train high-performance CLIP models with significantly reduced computations. Building upon this work, we hereby present CLIPA-v2 with two key contributions. Technically, we find this inverse scaling law is also applicable in the finetuning stage, enabling further reduction in computational needs. Empirically, we explore CLIPA at scale, extending the experiments up to the H/14 model with \app13B image-text pairs seen during training. 

Our results are exciting --- by only allocating a budget of \textdollar10,000, our CLIP model achieves an impressive zero-shot ImageNet accuracy of 81.1\%, surpassing the prior best CLIP model (from OpenCLIP, 80.1\%) by 1.0\% and meanwhile reducing the computational cost by $\app39\times$. Moreover, with an additional investment of \textdollar4,000, we can further elevate the zero-shot ImageNet accuracy to 81.8\%. 
\vspace{-1em}
\end{abstract}

%%%%%%%%% BODY TEXT
\vspace{-3.5em}
\section{Introduction}
CLIP \cite{radford2021clip} has emerged as the pioneering foundation model that bridges the gap between text and images, ushering computer vision research into the ``post-ImageNet'' era \cite{openclip,li2022flip,yuan2021florence,alayrac2022flamingo,ramesh2021dalle,rombach2022stable,schuhmann2021laion,xu2023cit,cui2022foundation}. However, the demanding computational requirements of CLIP hinder its widespread exploration. 
The recent work CLIPA \cite{li2023inverse} offers a computationally efficient solution --- with the introduction of an \textit{inverse scaling law} for CLIP training, it reveals that larger models can be trained with fewer input tokens. Building upon this observation, CLIPA demonstrates its efficacy in scenarios with limited computational resources, leading to a substantial reduction in the training cost of CLIP.

\begin{table*}[t]
    \centering
    \resizebox{0.78\linewidth}{!}{
    % \footnotesize
    \begin{tabular}{ccccc|cc}
       model &\# image token &\# text token  &data  source &\# seen samples &total compute ($\times 1e11$) &IN-1K \\
       \shline
        CLIPA-L/16 &36 &8 &LAION-400M &2.56B + 128M &0.5 &69.3 \\
        \hline
        \multirow{3}{*}{CLIPA H/14} & \multirow{3}{*}{36} &\multirow{3}{*}{8} &LAION-400M &2.56B + 128M &0.8 &72.8 \\
        & & &\textbf{LAION-2B} &2.56B + 128M &0.8 &74.1 \\
        & & &LAION-2B &\textbf{12.8B + 128M} &4 &\textbf{77.9} \\
    \end{tabular}
    }
    \vspace{-1em}
    \caption{\textbf{Scaling up CLIPA-v1~\cite{li2023inverse}.} Specifically, we explore scaling from the aspects of data, model, and schedule. We pretrain the H/14 model with 36 image tokens ($84\times84$) and 8 text tokens; for finetuning, we use 256 ($224\times224$) image tokens and 32 text tokens, following~\cite{li2023inverse}.}
    \vspace{-1em}
    \label{tab:scale_up}
\end{table*}

%--
This report provides a follow-up on CLIPA. Firstly, we  validate that the inverse scaling law is also applicable when finetuning models with input tokens at full resolution. This further reduces the training cost of CLIPA. 
Secondly, we investigate the performance of CLIPA at scale across various aspects, including model size (up to H/14), data (up to DataComp-1B~\cite{gadre2023datacomp} and LAION-2B~\cite{schuhmann2021laion} datasets), and training schedule (up to \app13B samples seen).

With these two contributions, we can train CLIP models with strong zero-shot performance on ImageNet \cite{deng2009imagenet}, meanwhile significantly reducing training costs. For instance, we can train a H/14 model with 81.1\% accuracy within a \textdollar10,000 budget. 
We stress that, compared to the best publicly available CLIP model from OpenCLIP \cite{openclip}, ours is both better (\textbf{+1.0\%}) and faster (by \app$\mathbf{39\times})$. Moreover, we can further boost this accuracy to 81.8\%, with an additional \textdollar4,000 investment.
These results are exciting as no prior work has thus far reached a similar performance within this small budget limitation. By open-sourcing our training code and models, we hope to contribute to the broader advancement and adoption of advanced CLIP models.

\begin{table*}[htbp]
  \begin{minipage}{0.28\textwidth}
    \centering
    \resizebox{\linewidth}{!}{
    \begin{tabular}{c|ccc}
       masking ratio &random &block  &grid \\
       \shline
       25\% &\textbf{78.2} &78.0  &77.9 \\
       50\% &\textbf{77.7} &77.6 &77.6 \\
       75\% &\textbf{76.2} &74.3  &\textbf{76.2} \\
    \end{tabular}
    }
    \vspace{-.87em}
    \caption{Comparison of different masking strategy. The results are obtained on on the LAION-2B dataset with H/14 model.}
    \vspace{-.65em}
    \label{tab:mask_strategy}
  \end{minipage}%
  \hfill
  \begin{minipage}{0.65\textwidth}
    \centering
    \resizebox{.9\linewidth}{!}{
    \begin{tabular}{c|cccccc}
       case &masking ratio &resolution  &\# seen samples &training FLOPs &IN-1K\\
       \shline
       CLIPA-v1 &{0\%} &{$224^2$} &128M &177.0G  &77.9 \\
        \hline
        (1) &30\% &{$224^2$}  &128M &135.9G    &78.0\\
        (2) &30\% &{$224^2$}  &512M &135.9G    &78.6\\
        (3) &30\% &{$224^2$}  &640M &135.9G    &78.5\\   
        (4) &40\% &{$336^2$} 
        &$640M$ &237.8G    &78.9\\
        \rowcolor{mygray} (5) &30\%+40\% &{$224^2+336^2$}   &512M+128M &156.3G   &\textbf{79.1}\\
    \end{tabular}
    }
    \vspace{-.87em}
    \caption{\textbf{Ablation of CLIPA-v2.} 
    In case (5), we use $224\times224$ input with a masking ratio of 30\% for the first 512M samples, and $336\times336$ input with a masking ratio of 40\% for the rest 128M samples.}
    \vspace{-.65em}
    \label{tab:ablation}
  \end{minipage}
\end{table*}

\section{Background}

CLIP has been a prominent foundation model due to its exceptional zero-shot capability and remarkable versatility ~\cite{radford2021clip,jia2021align}.
The tremendous success of CLIP can be attributed to the extensive scale of both the data \cite{radford2021clip,schuhmann2022laion5b,jia2021align,changpinyo2021cc15m,yuan2021florence,zhu2023c4data} and the model \cite{yu2022coca,openai2023gpt4,EVA-CLIP} it is built upon.
Nevertheless, it also poses a significant cost barrier to researchers who wish to train a strong CLIP model.
To reduce the computational burden, the recent work by Li et al. \cite{li2023inverse} presents an inverse scaling law, which reveals that larger models can effectively utilize fewer input tokens for training without severe performance drop, therefore enabling highly efficient CLIP training. As a byproduct of this discovery, the CLIPA models are introduced, which attain a zero-shot top-1 ImageNet accuracy of 69.3\% and can be trained on an 8 A100-GPU machine in just 4 days. 

Our work is built upon CLIPA~\cite{li2023inverse}, but focuses on furthering its efficiency and scaling it up.

\begingroup
\renewcommand{\arraystretch}{1.3}
\begin{table*}[t!]
     \centering
      \resizebox{\linewidth}{!}{
     \begin{tabular}{llllcr|cccccc|cccc}
     &  &  &  &  &  &\multicolumn{6}{c|}{\textbf{zero-shot classification}} &\multicolumn{4}{c}{\textbf{zero-shot retrieval}} \\
     &  &  &  &  &  &\multirow{2}{*}{IN-1K} &\multirow{2}{*}{IN-V2} &\multirow{2}{*}{IN-A} &\multirow{2}{*}{IN-R} &\multirow{2}{*}{ObjectNet} &\multirow{2}{*}{IN-SK} &\multicolumn{2}{c}{COCO} &\multicolumn{2}{c}{Flickr30k} \\
     &Models &Data Source  &\# seen samples@input size  &GPU hours\footnotemark[1]
     &Est. cost \footnotemark[2] &  &  &  &  &  &  &image  &text  &image  &text \\ \shline
     {OpenCLIP} &   &  &32.0B@$224^2$ &216,712 &\textdollar247,864  &78.0  &70.8  &59.2  &89.3  &69.7  &66.6   &49.5  &66.0  &77.8  &90.8 \\
     \rowcolor{mygray} \textbf{CLIPA-v2} &\multirow{-2}{*}{H/14}  &\multirow{-2}{*}{LAION-2B} &12.8B@$84^2$ + 512M@$224^2$ + 128M@$336^2$  &8,640 &\textdollar13,613 &79.1 &72.3  &71.7  &92.7  &69.9  &70.0   &50.2  &67.5  &78.2  &92.3 \\
    \hline
    \multirow{2}{*}{OpenCLIP} &L/14 &{DataComp-1B} &12.8B@$224^2$ &41,472 &\textdollar47,434 &79.2 &72.1  &69.6  &90.8  &74.3  &68.0   &45.7  &63.3  &73.4  &89.5  \\
    &G/14* &LAION-2B&32.0B@$224^2$ + 6.7B@$224^2$  &232,448 &\textdollar366,105 &80.1 &73.6  &69.4  &92.2  &73.0  &68.9   &51.4  &67.3  &79.6  &92.9  \\
    \rowcolor{mygray}\textbf{CLIPA-v2}  &H/14&{DataComp-1B}  &12.8B@$70^2$ + 512M@$224^2$  &5,920 &\textdollar9,324 &81.1 &74.7  &76.2  &93.7  &72.7  &72.4   &49.1  &67.1  &76.1  &92.4  \\
    \hline
    \rowcolor{mygray} &  &   &12.8B@$84^2$ + 512M@$224^2$ &4,008  &\textdollar6,318 &79.7 &72.8  &73.2  &92.1  &71.1  &69.3   &46.3  &64.1  &73.0  &89.1  \\
    \rowcolor{mygray} & \multirow{-2}{*}{L/14} &   & +128M{@$336^2$} &+512 &+\textdollar806 &80.3 &73.5  &77.7  &93.3  &73.1  &70.9   &47.2  &65.5  &74.6  &90.5  \\
    \rowcolor{mygray} &   &  &12.8B{@$84^2$} + 512M{@$224^2$} &7,776 &\textdollar12,247 &81.5 &75.0  &76.9  &94.3  &74.1  &72.7   &49.1  &67.0  &75.7  &90.6  \\
    \rowcolor{mygray} \multirow{-4}{*}{\textbf{CLIPA-v2}} & \multirow{-2}{*}{H/14} &\multirow{-4}{*}{DataComp-1B}  & +128M{@$336^2$}  &+864 &+\textdollar1,366 &\textbf{81.8} &\textbf{75.6}  &\textbf{82.7}  &\textbf{94.4}  &\textbf{77.4}  &\textbf{72.8}   &49.2  &67.2  &76.3  &90.3  \\
     \end{tabular}}
    \vspace{-1em}
    \caption{\textbf{Comparison with OpenCLIP~\cite{openclip}.} Our CLIPA-v2's GPU hour is estimated using an 8-A100 80GB GPU machine on Google Cloud, while the OpenCLIP's GPU hour is calculated based on their report$^1$. The corresponding training cost is estimated based on 80GB A100's cloud pricing$^2$. * denotes this model is trained with FLIP at a masking ratio of 50\%.}
    \label{tab:sota_comparison}
    \vspace{-1.15em}
\end{table*}
\endgroup

\begin{figure}[t!]
    \centering
    \includegraphics[width=0.75\linewidth]{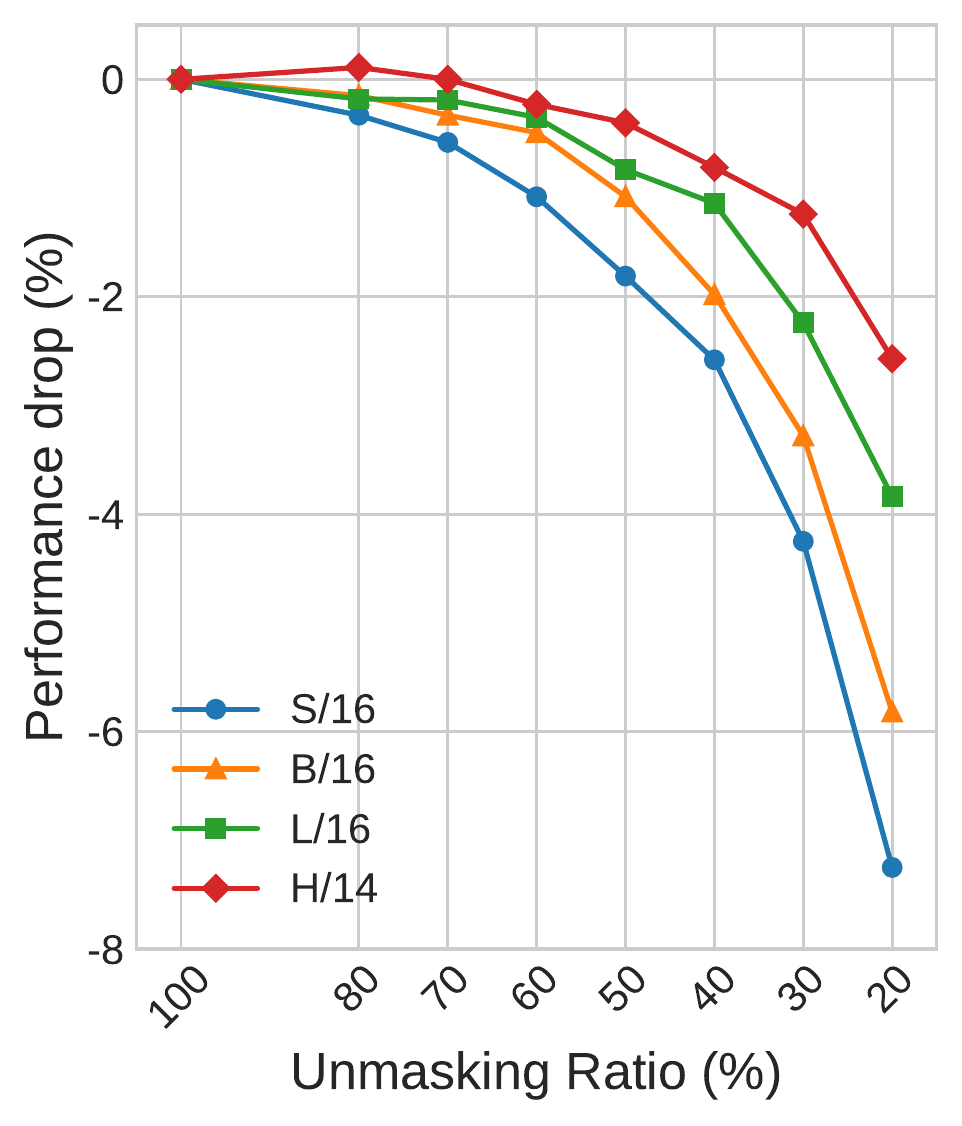}
    \vspace{-1em}
    \caption{\textbf{The \textit{inverse scaling law} on finetuning.} All models are finetuned with 128M samples, where we employ random masking for token reduction. 
    }
    \vspace{-1em}
    \label{fig:finetuning_inverse}
\end{figure}

\section{Experiments}
Our experiments contain three parts. Firstly, we check the applicability of \textit{inverse scaling law} during the finetuning stage with full-resolution tokens. 
Next, we scale up CLIPA in terms of data, model, and schedule. Lastly, we compare with other advanced CLIP models in terms of performance and computation cost. Our pretraining setup strictly follows CLIPA~\cite{li2023inverse}. We report the corresponding zero-shot top-1 accuracy on ImageNet~\cite{deng2009imagenet}. 

\paragraph{\textit{Inverse scaling law} in the finetuning stage.}
Following~\cite{li2023inverse},
we choose four different scales of models: S/16, B/16, L/16, and H/14, and train them on LAION-400M dataset. 
Random masking \cite{li2022flip,he2021masked} is used as the image token reduction strategy.
As shown in Figure~\ref{fig:finetuning_inverse}, larger models consistently exhibit a lower performance drop compared to smaller models when finetuning with the same number of input tokens. For instance, retaining 50\% of the input tokens merely results in a performance drop of 0.4\% for the H/14 model, compared to much higher drops of 0.8\% for L/16, 1.1\% for B/16, and 1.8\% for S/16. 

These results confirm the existence of the inverse scaling law in the finetuning stage, which enables us to reduce the required computations for CLIP training further.

\paragraph{Scaling up CLIPA~\cite{li2023inverse}.}
We next investigate the scaling behavior beyond the largest case studied in CLIPA. Specifically, our scaling efforts cover three aspects: model, data, and training schedule. The results are reported in Table~\ref{tab:scale_up}. 

First, we can observe that scaling the model size from L/14 to H/14 boosts the performance from 69.3\% to 72.8\%. Furthermore, we note switching the training dataset from LAION-400M \cite{schuhmann2021laion} to LAION-2B \cite{schuhmann2022laion5b} yields another 1.3\% improvement, suggesting the importance of data diversity. Lastly, by increasing the training schedule by a factor of 5, resulting in a total of \app13B seen samples, we achieve an impressive performance of 77.9\%. We stress that this scaled version of CLIPA H/14 model readily outperforms its counterpart in FLIP~\cite{li2022flip} by 0.3\% while requiring only $1/3$ of the training budget. 

These results confirm the efficiency and effectiveness of training CLIPA at scale. Next, we set this CLIPA H/14 with 77.9\% performance as our baseline for further ablation in the finetuning stage.

\paragraph{Ablation.}
In addition to random masking, we hereby investigate how grid masking and block masking affect finetuning performance. The results are reported in Table \ref{tab:mask_strategy}. Interestingly, compared to finetuning input tokens at the full resolution, we observe that 25\% masked random finetuning and block finetuning all lead to a slight performance improvement. With a larger masking ratio, all these masking strategies will lead to worse performance than full-resolution fine-tuning; but overall, random masking consistently yields stronger performance than the other two masking strategies.

We next ablate different finetuning setups and summarize the results in Table \ref{tab:ablation}. We choose 30\% masked random finetuning as the default strategy, as it leads to a slight performance improvement (+0.1\%) and enables a $1.3\times$ speedup of the finetuning process. Furthermore, adopting a $4\times$ finetuning schedule results in an additional improvement of 0.6\%. However, further increasing the finetuning schedule does not lead to any substantial performance gains.

Following~\cite{openclip}, we also investigate progressively finetuning with large image resolutions. Initially, for the first 512 million samples,  we finetune the model using a $224\times224$ input size with a masking ratio of 30\%; subsequently, for the remaining 128 million samples, we adopt a larger $336\times336$ input size with a masking ratio of 40\% and a smaller learning rate. As shown in the last row of Table \ref{tab:ablation}, \ie, case (5), progressive finetuning results in a slight performance improvement of 0.2\% compared to direct finetuning with a $336\times336$ input size and meanwhile achieving a notable $1.5\times$ speedup of the finetuning process.

\footnotetext[1]{We measure OpenCLIP~\cite{openclip}'s training time based on \url{https://laion.ai/blog/large-openclip/} and \url{https://laion.ai/blog/giant-openclip/}.
}

\footnotetext[2]{We estimate the total training cost based on \url{https://cloud.google.com/compute/gpus-pricing}, which is \textdollar1.575 per GPU hour, and \url{https://lambdalabs.com/service/gpu-cloud/pricing},  which is \textdollar1.5 per GPU hour.}

\paragraph{Comparison with OpenCLIP~\cite{openclip}.}
We summarize the results in Table \ref{tab:sota_comparison}.
Firstly, when trained on the LAION-2B dataset, our CLIPA-v2 H/14 model outperforms OpenCLIP's version by 1.1\% (79.1\% \vs 78.0\%) and meanwhile significantly reducing the training cost by $\app18\times$. Furthermore, when upgrading to the DataComp-1B dataset, our CLIPA-v2 H/14 (pretrained on images at $70\times70$) achieves an impressive zero-shot ImageNet accuracy of \textbf{81.1\%}, while keeping the training cost within \textdollar10,000. Notably, this 81.1\% accuracy is 1.0\% higher than the prior best CLIP model, which is OpenCLIP's G/14 model with a zero-shot ImageNet accuracy of 80.1\%. 

With an additional investment of  \textdollar4000, we can further enhance CLIPA-v2's training by 1) pretraining with a larger resolution (the image size from 70 to 84) and 2) applying the progressive finetuning with a larger image resolution of $336\time336$. These enhancements lead to an additional 0.7\% improvement, resulting in the \textit{best-performing CLIP model to date with an 81.8\% zero-shot ImageNet accuracy}. 

We also validate the superiority of CLIPA-v2 models on zero-shot robustness. For example, our 81.8\% H/14 model consistently yields much stronger performance than OpenCLIP's 80.1\% G/14 model on IN-V2 \cite{recht2019imagenetv2} (75.6\% \vs 73.6\%), IN-A \cite{hendrycks2021imageA} (82.7\% \vs 69.4\%), IN-R \cite{hendrycks2021imageR} (94.4\% \vs 92.2\%), ObjectNet \cite{barbu2019objectnet} (77.4\% \vs 73.0\%), and IN-SK \cite{wang2019imageS} (72.8\% \vs 68.9\%). However, we note that, when evaluating zero-shot retrieval performance on COCO \cite{lin2014coco} and Flickr30k \cite{plummer2015flickr30k}, OpenCLIP's 80.1\% G/14 model still performs the best. We conjecture this performance advantage should be attributed to the difference in training datasets, as Table \ref{tab:sota_comparison}'s results empirically suggest models trained with LAION-2B are better at retrieval tasks than models trained with DataComp-1B. 

We have open-sourced these advanced CLIP models in both JAX and PyTorch to facilitate future research.

\section*{Acknowledgement}
This work is supported by a gift from Open Philanthropy, TPU Research Cloud (TRC) program, and Google Cloud Research Credits program.

{\small
\bibliographystyle{ieee_fullname}
\bibliography{egbib}
}

\end{document}